# Machine Learning-Based Framework for Real Time Detection and Early Prediction of Control Valve Stiction in Industrial Control Systems

N. Promsricha[1,3], C. Chatpattanasiri[1,2], N. Kerdgongsup[3], S. Balabani[1,2]
[1]Department of Mechanical Engineering, University College London, UK
[2]Hawkes Institute, University College London, UK
[3]Thaioil Public Company Limited, Thailand

## ABSTRACT

Control valve stiction, a friction that prevents smooth valve movement, is a common fault in industrial process systems that causes instability, equipment wear, and higher maintenance costs. Many plants still operate with conventional valves that lack real time monitoring, making early predictions challenging. This study presents a machine learning (ML) framework for detecting and predicting stiction using only routinely collected process signals: the controller output (OP) from control systems and the process variable (PV), such as flow rate. Three deep learning models were developed and compared: a Convolutional Neural Network (CNN), a hybrid CNN with a Support Vector Machine (CNN-SVM), and a Long Short-Term Memory (LSTM) network. To train these models, a data-driven labeling method based on slope ratio analysis was applied to a real oil and gas refinery dataset. The LSTM model achieved the highest accuracy and was able to predict stiction up to four hours in advance. To the best of the author's knowledge, this is the first study to demonstrate ML based early prediction of control valve stiction from real industry data. The proposed framework can be integrated into existing control systems to support predictive maintenance, reduce downtime, and avoid unnecessary hardware replacement.

**Keyword:** valve stiction, process control, deep learning, CNN, CNN-SVM, LSTM, predictive maintenance

## 1. INTRODUCTION

In modern industrial process control systems, maintaining stable and efficient operation is essential for ensuring product quality, energy efficiency, and operational safety. One persistent challenge in these systems is control valve stiction, a form of static friction that prevents smooth valve movement when small changes occur in the controller output (OP) signal. Stiction leads to delayed or erratic valve behavior, resulting in process oscillations, actuator wear, and inefficient energy usage. These effects often degrade closed-loop control performance, especially in Proportional-Integral-Derivative (PID) controlled systems [1]. The controller may misinterpret the delayed valve response as a process disturbance, prompting unnecessary corrective actions that further exacerbate oscillatory behavior. Over time, this can reduce the overall efficiency of the control loop, increase maintenance costs, and affect long-term equipment reliability. These effects highlight the critical need for accurate detection and mitigation of stiction to ensure optimal process control and equipment longevity.

There are different types of control valves used in flow control applications, including globe, ball, butterfly, and gate valves, each suited for specific operational requirements. Ball and butterfly valves are often favored for on–off service and applications requiring low pressure drops, while gate valves are typically used where tight shutoff is necessary. In contrast, globe valves are designed for finer throttling and precise flow regulation, making them especially valuable in processes that demand continuous adjustment of flow rates. In large-scale and continuous industries such as petrochemical refining, power generation, and chemical processing, control valves play a central role in maintaining stable flows and pressures, ensuring uninterrupted operation and consistent product quality. Among these, globe valves are widely employed not only for their precise control capability but also for their reliability in handling varying process conditions, making them one of the most common valve types in advanced process control systems.

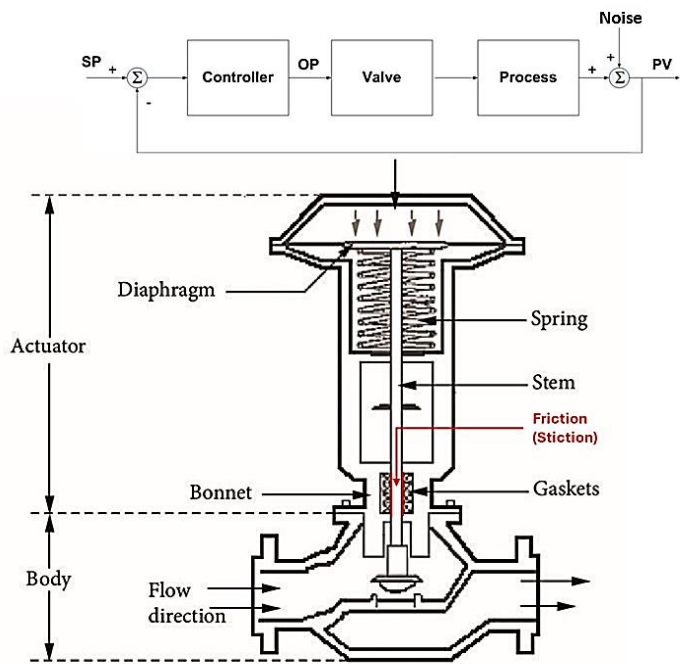 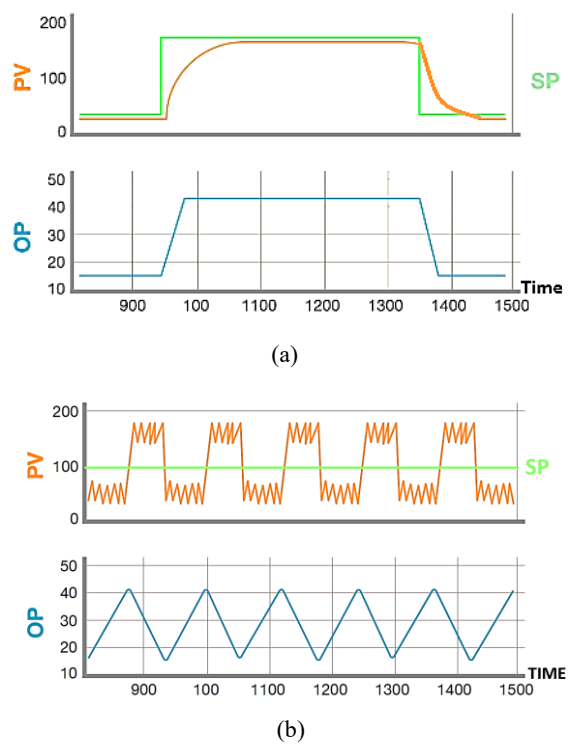

**Figure 1:** Control loop and Globe valve structure showing typical stiction location at the actuator stem packing [2].

**Figure 2:** Control valve operation graph showing (a) normal operation of a healthy control valve, (b) control valve operation with stiction behavior.

To better understand how globe valves operate and where stiction typically arises, their structure and control loop can be examined in more detail. Figure 1 illustrates a typical feedback control loop and the internal structure of a pneumatic globe control valve. The upper diagram shows the overall loop, consisting of a setpoint (SP), controller, actuator, process, and sensor. The lower diagram depicts the valve's internal structure, with the packing between the actuator stem and the valve body highlighted in red arrow to indicate the region where stiction typically occur. This localized friction prevents the valve plug from responding smoothly to changes in the OP, which in turn causes observable distortions in the process variable (PV) [2].

Building on this, Figure 2 presents examples of how these effects appear in practice. In Figure 2(a), the valve operates normally: the PV (orange line) closely follows the setpoint (SP, green line) with smooth adjustments in the OP (blue line). In Figure 2(b), the valve exhibits stiction behavior, characterized by a stick–slip cycle. The PV oscillates around the SP in a repetitive pattern, while the OP shows a sawtooth profile as it gradually changes until the static friction is overcome, causing a sudden valve movement. This repeating stick–slip motion is a clear indicator of stiction in control valves.

Although many modern oil and gas refinery plants use smart control valves with features like position feedback, self-diagnostics, and health monitoring, the majority still rely on conventional valves without these capabilities [5]. When integrated into control systems such as Distributed Control Systems (DCS), conventional valves typically provide only basic signals, which are OP and PV. Detecting stiction in such systems often depends on manual inspection, operator judgement, or scheduled loop tuning, all of which are labor-intensive, and unsuitable for continuous monitoring or predictive maintenance [4], [6].

This thesis proposes a machine learning (ML) framework for real-time detection and early prediction of control valve stiction. Different deep learning methodologies are developed and applied on actual industrial data with the view to enhance early fault identification and predictive maintenance strategies.

# 2. LITERATURE REVIEW

In literature, control valve stiction has been studied using model-based diagnostics, light-weight regression and rule-based screening, and more recently deep learning. Model-based methods use a mathematical or physical representation of the control loop to predict normal behavior, then compare it against actual process data to identify deviations caused by stiction. They work well when loop dynamics are well known but require tuning and expert intervention. Regression and rule-based approaches are simple and fast yet struggle with temporal patterns. Convolutional Neural Network (CNN) deep models learn discriminative features for detection on OP and PV but rarely address forecasting. Long Short-Term Memory (LSTM) models excel at forecasting in other industrial domains, yet few studies apply them to valve stiction using real plant data.

## *2.1 Traditional and Model-Based Methods*

Patwardhan et al. [1], were among the first to propose a software sensor–based technique for detecting valve stiction through feedback loop analysis. Although effective in identifying oscillatory behavior, their method was sensitive to loop dynamics and required manual tuning. A different perspective was offered by Gopaluni and Larimore [4] , who applied multiscale Principal Component Analysis (PCA) to separate valve friction effects from normal process variations and noise. In a similar vein, Huang et al. [6] focused on analyzing limit cycles in control loops to estimate stiction magnitude, using a model-based detection and quantification framework. Telen et al. [5] provided a comparative assessment of model-based and data-driven techniques, noting that both can be effective but that model-based methods often demand substantial domain expertise. Overall, such techniques perform best in stable, well-characterized systems, but they remain largely reactive and require periodic tuning, which limits their suitability for continuous large-scale monitoring.

## *2.2 Regression and Rule-Based Techniques*

Instead of complex models, some researchers have explored simpler regression-based solutions. Damarla et al. [7] used linear regression to estimate the slope between OP and PV during transitions, enabling detection and quantification of stiction. Yazdi et al. [2] advanced this approach by integrating support vector machines (SVM) with a generalized statistical variable and slope-aware regression. This not only improves robustness against noise and adaptability to nonlinear data but also provides a detailed visualization of stiction location and severity. While these approaches are computationally efficient and easy to interpret, they lack the capacity to capture intricate time-series patterns and are not inherently suited for multi-step prediction tasks.

## *2.3 CNN for Stiction Detection*

Deep learning methods, particularly CNN, have shown strong capability in recognizing industrial signal patterns. For example, Amiruddin et al. [13] applied a multilayer perceptron (MLP) to OP and PV signals and achieved 78% detection accuracy on benchmark datasets. Building on this, Henry et al. [8] adapted an AlexNet-based CNN for simulated valve signals, reporting accuracy above 90%. Gunnell et al. [9] extended CNN applications to real-world signals using continuous wavelet transform (CWT) preprocessing, resulting in 75.76% accuracy. Hybrid models have also been explored. Henry et al. [10] combined CNN with PCA to reduce dimensionality and improve generalization, while Memarian et al. [12] leveraged Markov transition fields to enhance CNN feature extraction for valve fault classification. Navada et al. [11] incorporated ANN models with Kalman filtering, achieving over 76% accuracy in industrial loops. Although CNN-based methods excel at feature extraction for classification, their adoption for forecasting future stiction events remains limited.

## *2.4 LSTM and Forecasting Applications*

LSTM networks are designed to capture long-term dependencies in sequential data, making them suitable for fault prediction. Malhotra et al. [14] demonstrated this by applying an LSTM encoder–decoder to multivariate industrial datasets, achieving 83% anomaly detection accuracy. In another study, Filonov et al. [15] applied LSTM models to process sensor fault prediction with an 87% success rate. Predictive applications have also been reported in rotating machinery. Shokouhmand et al. [16] used LSTM to forecast compressor failures from vibration data, achieving 93% accuracy, while Zhao et al. [17] reached 97.1% accuracy in predicting bearing faults. These studies underscore LSTM's potential in moving from reactive detection to proactive maintenance strategies. However, to the best of the author's knowledge, LSTM has not been applied to studies related to real-time or early detection of control valve stiction.

## 2.5 Summary of Key Findings and Research Gap

A summary of the key literature reviewed is presented in Table 1. The table highlights that prior studies emphasize either offline detection or simulated signals, with few using real OP and PV from plant historians, which are systems that store and organize large amounts of sensor and control data from industrial plants for long-term analysis. Almost none compare detection and multi-step early prediction under consistent datasets and window configurations. A direct benchmark of CNN-based detection against LSTM-based forecasting on the same industrial refinery data is missing.

**Table 1:** Summary of existing ML methods for control valve stiction detection and time-series fault prediction.

| Reference | Method | Input Type | Reported Accuracy | Real Time Detection | Early Prediction | Equipment | Data Type Used |
|---|---|---|---|---|---|---|---|
| Amiruddin et al., 2019) [13] | Multilayer Perceptron (MLP) with one layer | OP, PV | 78% | ✓ | ✗ | Control Valve | Real Industrial Data |
| Henry et al. (2021) [8] | CNN–AlexNet | OP, PV | 90% | ✓ | ✗ | Control Valve | Simulated Data |
| Gunnell et al. (2024) [9] | CWT–CNN | OP, PV | 75.76% | ✓ | ✗ | Control Valve | Real Industrial Data |
| Henry et al. (2020) [10] | CNN–PCA | OP, PV | 70.50% | ✓ | ✗ | Control Valve | Real Industrial Data |
| Navada et al. (2024) [11] | ANN + Kalman | OP, PV | 76% | ✓ | ✗ | Control Valve | Real Industrial Data |
| Memarian et al. (2023) [12] | MTF–CNN | OP, PV | 62.16% | ✓ | ✗ | Control Valve | Real Industrial Data |
| Malhotra et al. (2016) [14] | LSTM Encoder–Decoder | Multisensory time series | 83% | ✗ | ✓ | Process Sensors | Fault Prediction |
| Filonov et al. (2016) [15] | LSTM | Industrial sensor streams | 87% | ✗ | ✓ | Process Sensors | Fault Prediction |
| Shokouhmand et al. (2020) [16] | LSTM | Vibration signals | 93% | ✗ | ✓ | Compressor | Fault Prediction |
| Zhao et al. (2020) [17] | LSTM | Bearing vibration sensor data | 97.13% | ✗ | ✓ | Bearing | Fault Prediction |

✓ indicates that the information is presented in the research, ✗ indicates that the information is not presented in the research

## 2.6 Aims and Objectives of This Study

This research aims to develop and evaluate a ML framework for detecting and predicting control valve stiction using real industrial process data. The approach combines real-time detection and multi-step early prediction within a unified setup. Prior studies on deep learning for stiction detection report accuracies of 70 percent or higher. Accordingly, this work adopts 70 percent as the minimum acceptable accuracy to ensure practical reliability while reducing false alarms. To achieve this aim, the study focuses on three key objectives:

Objective-1: Collect OP and PV time-series data from an operational oil and gas refinery and apply a systematic method to label stiction events.

Objective-2: Develop real-time stiction detection models using deep learning approaches, namely CNN, a CNN with Support Vector Machine classifier (CNN–SVM), and LSTM. Compare their performance using the same labelled dataset.

Objective-3: Design and evaluate an early detection of stiction framework for multi-step prediction using matched input and lookahead window configurations, ensuring the performance threshold established for practical deployment is met to enable reliable predictive maintenance.

## 3. METHODOLOGY

The overall methodology is illustrated in Figure 3, which outlines the complete workflow from data acquisition to model evaluation. The process begins with the collection of OP and PV data from the Plant Information (PI) system. This is followed by preprocessing and the systematic labeling of stiction and non-stiction periods.

The prepared dataset is then used to develop deep learning models for two main tasks. Objective 2 focuses on designing real-time stiction detection models using CNN, CNN–SVM, and LSTM. Objective 3 focuses on developing early prediction models for multi-step prediction using the same architecture and matched window configurations.

For both tasks, each model's performance is evaluated independently to identify the most effective approach. The model that meets or exceeds the target accuracy threshold of 70 percent is selected as optimal, ensuring practical reliability for predictive maintenance and the capability to predict future stiction in industrial settings.

The next subsection describes the data processing workflow in detail, including preprocessing steps and the automatic labeling of stiction events.

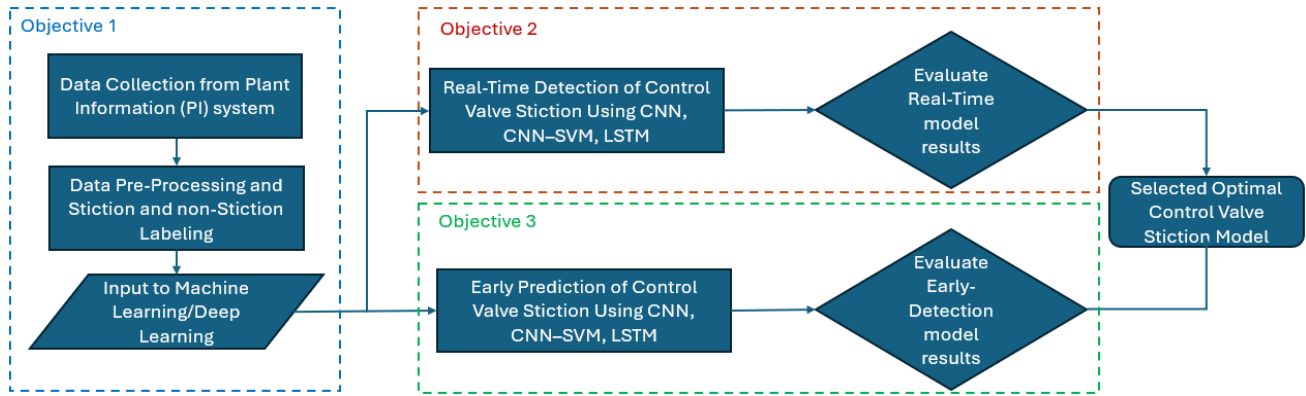

**Figure 3:** Flow chart of the research workflow, with each stage linked to its corresponding objective.

### 3.1 Dataset Collection and Preprocessing

The operational data used in this research was obtained from a flow control valve within an oil and gas refinery. Data acquisition is managed through the refinery's PI System, a process historian developed OSIsoft (OSIsoft LLC, CA, USA) [18] that stores processes and provides secure access to high-frequency time-series data from industrial equipment. The PI System was selected as the data source instead of direct access to Distributed Control Systems (DCS) or Programmable Logic Controllers (PLC) because it offers centralized, long-term historical storage and integrates data from multiple control systems, ensuring consistent accessibility without interfering with real-time plant operations [18].

The network architecture of the PI System is shown in Figure 4, following the ISA-95 hierarchy levels. At Level 0, the control valve operates within the process plant, with position and flow-related measurements captured by sensors. These signals are transmitted to Level 1 controllers such as DCS or PLCs, which handle immediate process control. At Level 2, PI Interfaces and Connectors gather data from controllers. The information is archived at Level 3 in the PI Data Archive for long-term storage.

A Demilitarized Zone (DMZ) at Level 3.5 enables secure access to PI Vision and the PI Web API without exposing the control network, before the data becomes available at Level 4 for business and analytical applications. While this multi-layered architecture ensures data integrity, cybersecurity, and system reliability, it also introduces challenges in data collection, requiring careful coordination and secure protocols to access the required datasets for analysis.

Over the span of one year, approximately 525,600 data points were recorded for both OP and PV signals, logged at a one-minute sampling interval by the PI system. The dataset includes intervals of both normal operation and suspected stiction behavior, providing a diverse and representative basis for supervised ML model training and evaluation.

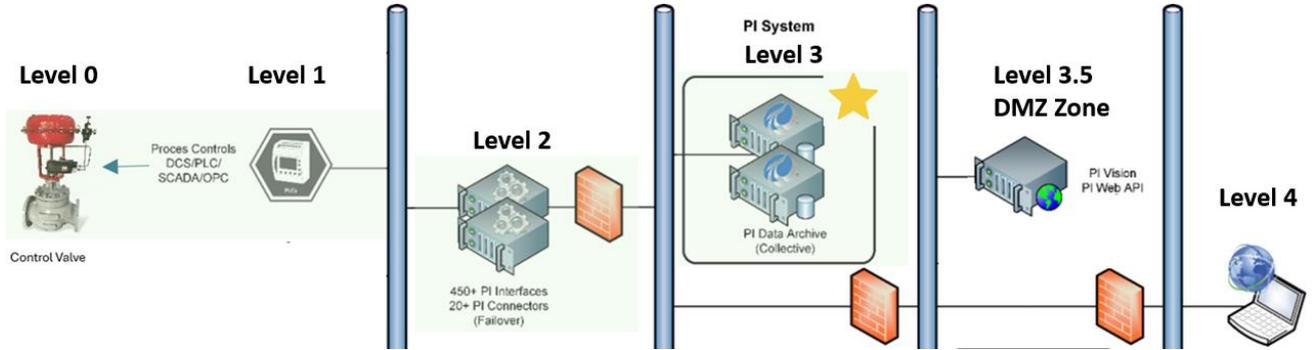

**Figure 4:** PI System architecture green highlighted showing data acquisition and flow from the control network to end-user systems. Adapted from OSIsoft Manual[18].

However, the raw exports from the PI system contained format inconsistencies and occasionally missing data. These gaps often occur because the PI System uses a compression algorithm that omits storing a new value if it has not changed from the previous timestamp, in order to save server storage space. To standardize the dataset, a custom Python preprocessing script was developed to replicate the PI system's behavior, ensuring consistent formatting and temporal alignment across all data points. The key steps included:
1. Date-time parsing and standardization for OP and PV time columns, handling diverse date formats.
2. Generation of a unified 1-minute interval time axis to align all values.
3. Filling missing values by carrying the last observation forward, with backward filling applied only at the start of the dataset to handle leading gaps.
4. Renaming and merging OP and PV into a unified data frame for model input.

Thus, the processed signals are:

$$\hat{V}(t_j) = \begin{cases} V_{(t_j)}, & \text{if } t_j \text{ exists in raw data} \\ V_{(t_{j-1})} \text{ or } V_{(t_{j+1})}, & \text{if } V_{(t_j)} \text{ is missing} \end{cases} \quad (1)$$

Where $t_i$ be the original timestamp of record $i$, $V_{op}(t_i)$ and $V_{pv}(t_i)$ be the corresponding OP and PV values, $t_j \in T$, where $T = \{t_{min}, t_{min} + 1\,min, \ldots, t_{max}\}$, be the uniform time axis.

This process ensures that each OP–PV pair is temporally aligned per minute, preserving the behavior fidelity of the original PI system. The success of this interpolation can be visualized in Figure 5, which compares the time-series plots from PI system and the processed Python output.

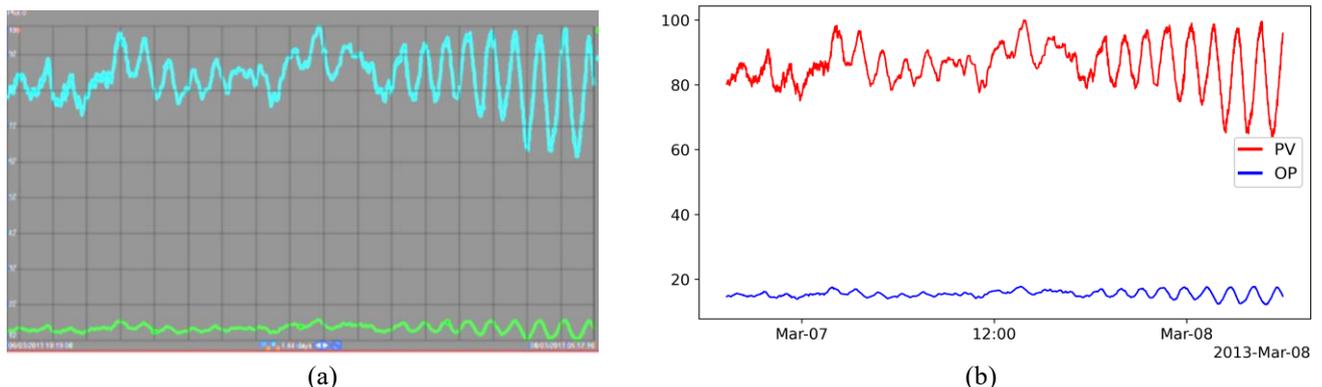

(a)           (b)

**Figure 5:** PV and OP plot comparison showing (a) data visualized directly from the PI system and (b) the same dataset processed and plotted in Python.

## 3.2 Data Labelling Method of Control Valve Stiction / Non-stiction for Model Input.

Based on the literature review, two main approaches are generally used to label stiction in control valves, first is model-based techniques, which require detailed process models and expert calibration, and second is data-driven techniques, which extract patterns directly from operational data. This study adopts a data-driven approach due to its suitability for large-scale application without the need for complex process modelling or extensive expert manual intervention.

The Slope Ratio Method, originally proposed by Damarla et al. [7], is used in this research to generate control valve stiction labels directly from process signals, avoiding the need for manual annotation by experts.

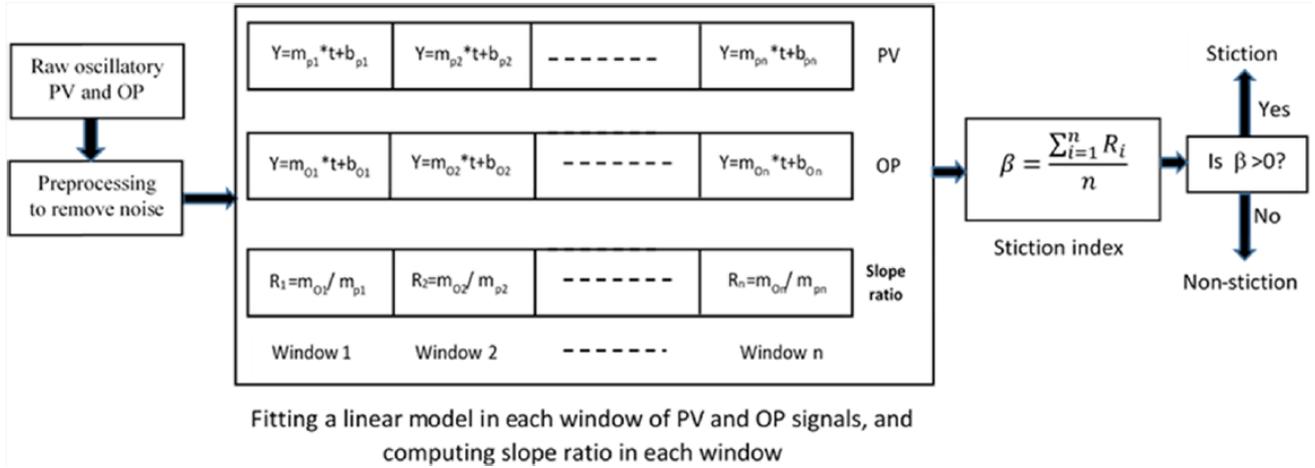

**Figure 6:** Slope Ratio Method framework for stiction classification. Adapted from Damarla et al. [7]

As illustrated in Figure 6, the method begins by dividing the time-series signals of the PV and OP into fixed-length windows of 60 minutes. Within each window, linear regression is applied to both PV and OP, producing equations in the form:

$$Y = m_{pv,i} * t + b_{pv,i} \tag{2a}$$

$$Y = m_{op,i} * t + b_{op,i} \tag{2b}$$

where, $m$ represents the slope, and $b$ is the intercept. The slope ratio $R$ for each window is calculated as:

$$R_i = \frac{m_{op,i}}{m_{pv,i}} \tag{3}$$

The stiction index $\beta$ is then computed as the average slope ratio over $n$ consecutive windows:

$$\beta = \frac{1}{n}\sum_{i=1}^{n} R_i \tag{4}$$

The stiction labels are assigned according to the value of $\beta$. If $\beta > 0$, the window is classified as stiction, whereas if $\beta \leq 0$, it is classified as non-stiction. This threshold follows Damarla et al. [7], where a positive average slope ratio indicates asymmetric valve response, a key indicator of stiction. While Damarla et al. evaluated $n$ values between 1 to 24 and identified $n = 15$ as optimal, this study also examined different n configurations using industrial control valve data to assess their impact on labeling performance.

The Hotelling's $T^2$ statistic was used as the second approach for training data generation. This multivariate statistical method captures the variance and correlation between the OP and PV signals to model normal operating behavior. Deviations from this model indicate abnormal conditions such as stiction. Anomalies are flagged when $T^2$ values exceed a statistical threshold, with the confidence percentile influencing detection sensitivity. While Yazdi et al. [2], considered 90% and 60% confidence percentiles, this study also evaluated intermediate values (e.g., 70% and 80%) as well as lower levels below 50% to ensure robustness.

The $T^2$ statistic was computed over fixed-length windows, where each window consisted of 60-minute segments of synchronized OP and PV data. For each window ($w_i$), the Hotelling's $T^2$ score was calculated as:

$$T^2(w_i) = (x_i - \mu)^T \sum{}^{-1} (x_i - \mu) \tag{5}$$

where $x_i$ represents the feature vector containing statistical summaries of OP and PV within window $w_i$, $\mu$ is the overall mean vector across all windows, $\Sigma^{-1}$ denotes the inverse of the covariance matrix of the features.

To determine whether a window reflects stiction behavior, a thresholding strategy was applied. The threshold was varied by percentile of the entire distribution of $T^2$ scores ($p$):

$$\text{Threshold}_{T^2} = \text{percentile}(T^2, p) \tag{6}$$

Any window with a $T^2$ scores greater than this threshold was labeled as stiction:

$$\text{Label}(w_i) = \begin{cases} 1 & \text{if } T^2(w_i) > \text{Threshold}_{T^2} \\ 0 & \text{otherwise} \end{cases} \tag{7}$$

The Slope Ratio Method and Hotelling's $T^2$ statistic were applied independently to generate stiction and non-stiction labels. The performance of each labeling approach will be compared with the better-performing method selected as the label for training and evaluating the ML models described in the next section.

### 3.3 Real-Time Stiction Detection using ML

Following the preparation of training data labels using the slope ratio and Hotelling's $T^2$ methods, the first major task of this research, in line with the second objective, was to develop ML models capable of real time detection of control valve stiction. This section describes the detection setup, configuration of the detection window, model architectures applied, and evaluation criteria used to assess performance.

#### 3.3.1 Detection Window and Common Parameters Configuration

For real time detection, the continuous time series data was divided into fixed length sliding windows. Each detection window represents a recent segment of process activity used to evaluate whether stiction is occurring. In this study, the window size was set to 60 minutes, corresponding to 60 data points, matching the input window size used for ground truth labeling. The sliding step was also set to 60 minutes to align detection intervals with the labeling process. This configuration reflects real world monitoring, where detection relies solely on historical and current data available at the time, without access to future measurements.

The training settings for both real-time detection and early prediction in Table 2 were selected based on established practices in time-series deep learning for process fault detection and supported by prior research on control valve stiction classification [8], [9], [14], [17], [29].

**Table 2:** Common Model Training Configuration

| Parameter | Value |
|---|---|
| Maximum Epochs | 100 |
| Learning Rate | 0.001 |
| Early Stopping Patience | 3 |
| Batch Size | 64 |
| Optimizer | Adam |
| Loss Function | Binary Cross Entropy |
| Training: Validation: Testing | 60:20:20 |
| Input Shape (Window) | 24 x 2 (PV, OP) |
| LSTM layer activation function | tanh |
| Pooling layer activation function | Rectified Linear Unit (ReLU) |

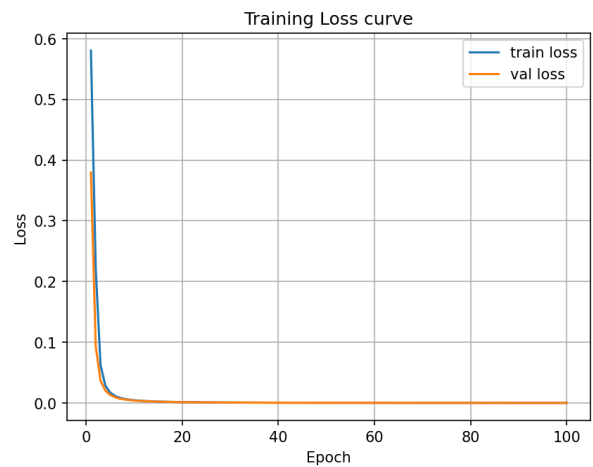

**Figure 7:** Example of Train/Validation loss in Model Training

An epoch refers to one complete pass through the entire training dataset. In this study, Early stopping was applied, such that training was halted if no improvement in the validation loss was observed for 3 consecutive epochs, thereby preventing overfitting and unnecessary computation, which allowed the loss values to stabilize without overfitting, consistent with recommendations in deep learning literature [29]. Figure 7 illustrates an example of training and validation loss curves across these epochs. Both curves show a rapid decline during the initial epochs, followed by a plateau, indicating that the models quickly learned relevant features and then maintained stable performance. The close alignment between the two curves suggests minimal overfitting and good generalization to unseen data.

The learning rate was set to 0.001, a value frequently used in similar fault detection tasks [14], [17], providing a stable balance between convergence speed and training stability. The Adam optimizer was chosen for its ability to adapt the learning rate dynamically, improving efficiency in training non-linear time-series models [29].

A batch size of 64 was used, meaning 64 samples were processed before the model parameters were updated. This choice, also found in similar LSTM and CNN applications [14], [17], offers a good compromise between computational efficiency and convergence stability. The binary cross-entropy loss function was applied since the task is a binary classification (stiction or no stiction).

The input shape (24 × 2) corresponds to the 60-minute base window with 24 consecutive readings for two signals: PV and OP. This configuration was determined in Section 4.1 as the most reliable labeling approach based on slope ratio analysis.

Activation functions followed common best practices for sequential data: tanh for LSTM layers to effectively handle time-dependent patterns and Rectified Linear Unit (ReLU) for pooling layers to accelerate training [29]. A training–validation–testing split of 60:20:20 was applied, ensuring that a portion of the data was withheld during training to evaluate generalization performance.

All model training and evaluation were performed on a laptop equipped with an Intel Core i7-13620H processor (6 performance cores and 4 efficiency cores), 64 GB DDR5-5600 RAM, and Windows 11 Home (64-bit). This hardware configuration provided sufficient computational capacity to support the deep learning workloads, enabling fast training iterations and smooth execution without the need for external high-performance computing resources.

### 3.3.2 Deep Learning Models for Real-Time Stiction Detection

Three deep learning architectures were selected for the real-time classification task: CNN, CNN–SVM, and LSTM. The choice was informed by the literature review, which highlighted CNN's strength in extracting discriminative features from OP and PV signals, the potential of CNN–SVM hybrids to improve decision boundaries in noisy industrial data, and LSTM's ability to capture temporal dependencies for predictive maintenance. For consistency and comparability, all three architectures were trained using the configuration described in Section 3.3.1, ensuring that performance differences reflected model design rather than training conditions [19].

CNN based stiction detection uses a one-dimensional CNN (1D CNN) to classify whether a given segment of OP and PV time-series data contains valve stiction. CNNs are effective at identifying short, local patterns such as oscillations, which often indicate stiction events. As shown in Figure 8, the labelled data segment passes through three convolutional layers with 128, 64, and 32 filters, each using a kernel size of 3. This kernel size is a common choice in time-series analysis because it captures short-term variations without over-smoothing the data [19]. The number of filters follows the power-of-two convention for computational efficiency and gradually decreases to focus on more abstract features in later layers.

The extracted patterns are flattened into a single vector and passed to two fully connected dense layers with 64 and 32 units, which combine the learned features while limiting overfitting. These dense layer sizes are a typical configuration in most related research [29] as a balance between complexity and generalization. The final output layer applies a sigmoid activation to produce a probability between 0 and 1, which is then converted into a binary decision: 0 for non-stiction and 1 for stiction.

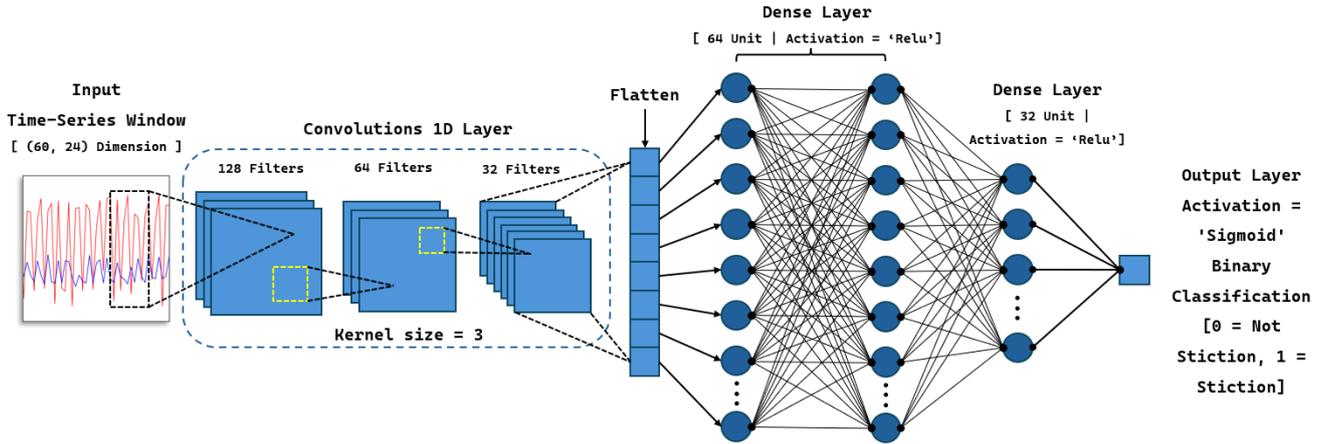

**Figure 8:** CNN architecture for stiction detection using 1D convolutions and dense layers for binary classification of time-series input.

The hybrid CNN-SVM based stiction detection model combines the feature extraction capabilities of CNN with the robust classification performance of SVM. CNN handles the initial pattern recognition from OP and PV signals, while SVM provides a more precise decision boundary in noisy industrial environments. As shown in Figure 9, the model first processes the input through the same convolutional feature extractor used in the CNN model, generating a set of flattened features that capture oscillation characteristics linked to stiction.

Instead of passing these features to dense layers for classification, the dense layers are connected to an SVM. The SVM with a radial basis function (RBF) kernel then performs the classification task. The RBF kernel is chosen because it can capture complex, non-linear relationships in the data, allowing more effective separation of stiction and non-stiction cases, even when the classes overlap in feature space [2].

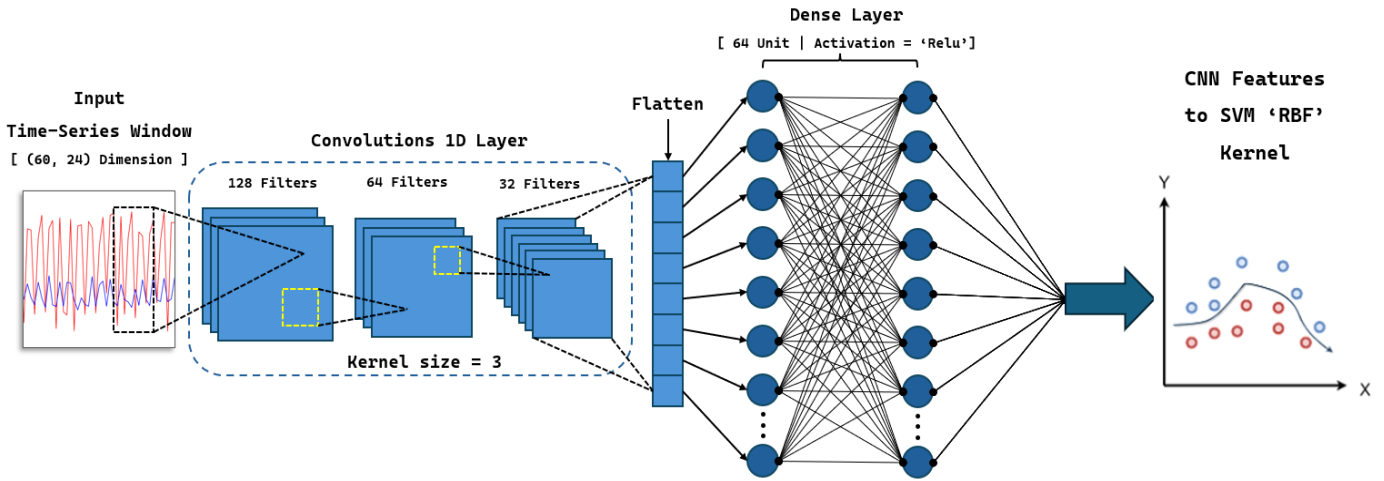

**Figure 9:** CNN-SVM hybrid model for stiction detection, where CNN-extracted features are classified using an SVM with an RBF kernel.

LSTM-based stiction detection leverages LSTM networks to capture temporal dependencies in OP and PV time-series data. Unlike CNNs, which excel at short local patterns, LSTMs preserve information across longer horizons, making them effective for detecting stiction trends that develop gradually. At the core of an LSTM is a memory cell ($C_t$), regulated by three gates: the forgetting gate decides how much past information to retain, the input gate determines which new information is added, and the output gate ($o_t$) controls how much of the updated memory is revealed. The candidate state $\tilde{C}_t$ provides new content, and together these elements update the memory according to:

$$C_t = f_t \odot C_{t-1} + i_t \odot \tilde{C}_t, \qquad h_t = o_t \odot tanh(C_t) \qquad (5)$$

where $h_t$ is the hidden state shown in Figure 10, representing the short-term output at time step $t$. The $\odot$ denotes element-wise multiplication, meaning each component of the gate output scales the corresponding element of the memory.

As illustrated in Figure 10, the proposed model processes a pre-processed time-series window through two stacked LSTM layers. The first layer 64 units, return sequence enabled outputs the full hidden state sequence for the second layer 32 units, which condenses the temporal patterns. A dense layer with 32 units and ReLU activation further integrates these features, followed by a sigmoid output neuron that classifies the sequence as stiction (1) or non-stiction (0). This configuration balances temporal depth with computational efficiency, enabling the model to detect both rapid oscillations and slow operational drifts in real time [14], [16], [17].

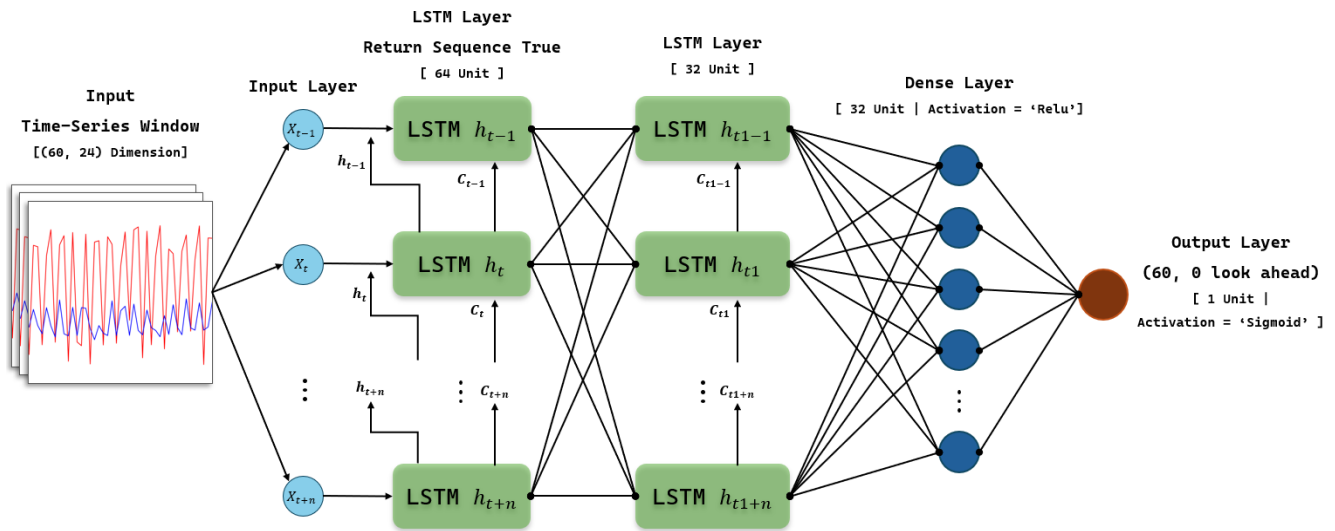

**Figure 10:** LSTM-based model for stiction detection, using two stacked LSTM layers to learn temporal patterns from time-series input for binary classification.

### *3.4 A Framework for Early Prediction of Valve Stiction*

This section investigates the potential of early prediction of control valve stiction, aligning with the third research objective, by leveraging three deep learning models, namely CNN, CNN-SVM, and LSTM, under varying detection and forecasting window configurations. To simulate real-world conditions where early warning is essential, time-series data was segmented into two continuous windows: a detect window (used as input for learning) and a lookahead window (used for forecasting stiction). Each window represents 60 minutes of process data. Both detect and lookahead windows were varied from 1 to 4 hours, generating multiple input–output combinations (e.g., 1–1, 1–2, 1–3, ..., 4–4) to evaluate optimal configurations. This range was selected to provide a balance between short-term predictive accuracy and sufficient early warning time for operator intervention, while keeping computational demands feasible for the prototype developed in this study and aligned with the overall project timeframe.

As shown in Figure 11(a), the labeling process begins by selecting a detect window containing 60 minutes of OP and PV time-series data. This detect window is immediately followed by a lookahead window of equal or varying length. If stiction is detected within the lookahead window (highlighted area in the waveform), the preceding detect window is assigned a positive label (Label 1). If no stiction occurs, it is assigned a negative label (Label 0). This creates a labeled dataset where the model can associate specific temporal patterns in the detect window with future stiction events

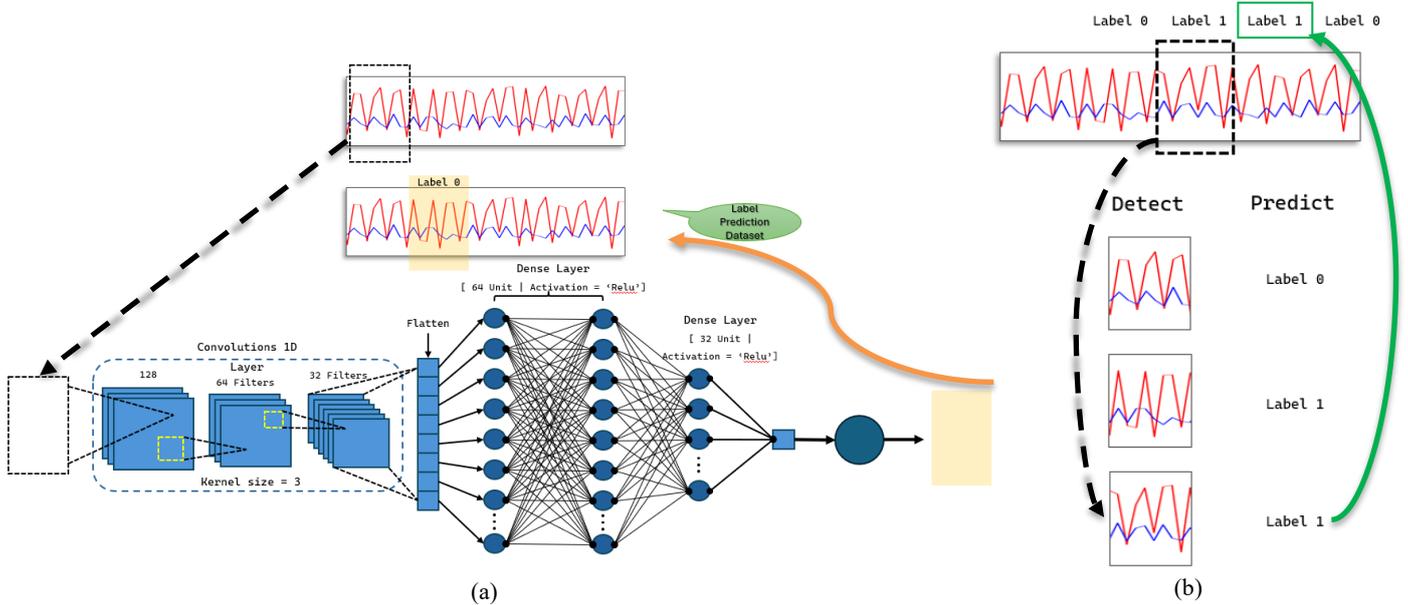

**Figure 11:** (a) Early prediction window labeling workflow, showing how time-series segments are assigned labels for training, and (b) model prediction process for detected patterns with one-window lookahead configuration.

Figure 11(b) illustrates how the trained model uses this labeling framework to make predictions. The detect window is provided as input to the model, which processes it through ML model before producing an output label. The prediction indicates whether stiction is expected in the upcoming lookahead window. This framework effectively transforms the model into an early warning system, capable of identifying signs of valve stiction before they occur.

The full set of detect–lookahead window configurations used in these experiments is provided in Appendix 9.2. These configurations range from 1 to 4 hours for both detect and lookahead windows, with each pairing representing a distinct experimental setup. The appendix also includes visual diagrams that show how each combination segments the data, where blue highlights the detect window and orange highlights the corresponding lookahead window.

### 3.5 Evaluation Metrics for Real-Time and Early Prediction

To assess the real-time and early prediction performance of the proposed deep learning models for stiction detection, standard binary classification metrics were employed. These include the confusion matrix, which visualizes the alignment between actual and predicted labels (Figure 12), and the classification report, which summarizes key performance indicators such as accuracy, precision, recall, and F1-score, as presented in Table 3. In addition to these metrics, a heatmap was employed to present early prediction performance across various detect–lookahead window combinations.

**Table 3:** Evaluation metrics: definitions, formulae and interpretations.

| METRIC | FORMULA | INTERPRETATION |
|---|---|---|
| ACCURACY | $\dfrac{TP + TN}{TP + TN + FP + FN}$ | Overall correctness of the model across all predictions |
| PRECISION | $\dfrac{TP}{TP + FP}$ | Proportion of predicted stiction instances that were actually stiction |
| RECALL | $\dfrac{TP}{TP + FN}$ | Proportion of actual stiction instances that the model correctly detected |
| F1-SCORE | $2 \times \dfrac{Precision \times Recall}{Precision + Recall}$ | Harmonic mean of precision and recall; balances false positives and false negatives |

Where TP = True Positive, TN = True Negative, FP = False Positive, FN = False Negative

In this study, the confusion matrix terms are defined in the context of valve stiction prediction within the lookahead window, as follows:

True Positive (TP): Instances where the model correctly predicts stiction in the lookahead window when stiction is actually present.

False Positive (FP): Instances where the model predicts stiction in the lookahead window, but no stiction occurs in reality.

True Negative (TN): Instances where the model correctly predicts no stiction in the lookahead window when there is indeed no stiction.

False Negative (FN): Instances where the model fails to predict stiction in the lookahead window, even though stiction is actually present.

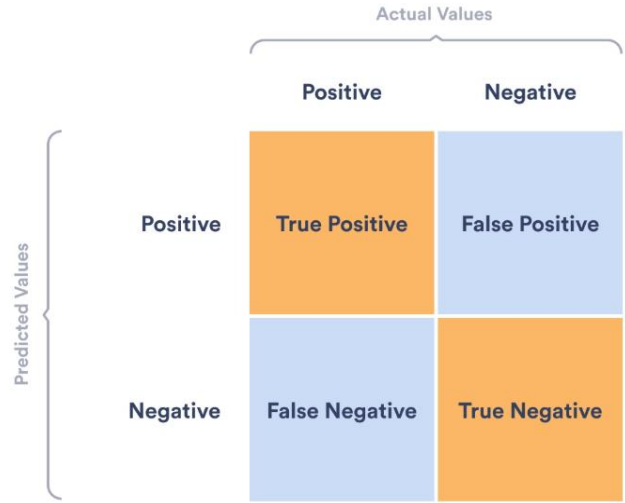

**Figure 12:** Typical Confusion Matrix illustrating the performance of the ML models [20]

These metrics offer valuable insights into each model's discrimination capability. This is particularly important in industrial environments, where a high number of false negatives may lead to missed fault detection and potential system failures, while a high number of false positives could result in unnecessary maintenance actions. Both scenarios can negatively impact system reliability, safety, and operational efficiency.

# 4. RESULTS

This section reports a comparison of the two automatic labeling approaches to obtaining reliable training labels, followed by an evaluation of the detection and early prediction performance of the CNN, the hybrid CNN-SVM, and the LSTM models under the window configurations defined in Section 3.

## *4.1 Comparison of Training Data Labeling Approaches*

Two automatic labeling methods were examined: the slope ratio method and Hotelling $T^2$. The slope ratio method uses linear regression over fixed-length windows to capture asymmetric valve behavior, while the Hotelling $T^2$ method applies a statistical threshold to detect deviations from normal operating conditions. Previous research identified an optimal configuration for the slope ratio method; in this study, both methods were re-evaluated using industrial control valve data.

Figure 13(a) illustrates the results when $n = 15$ consecutive windows are used. The red-highlighted bands represent detected stiction periods. This configuration produces fragmented and intermittent labeling, where only partial sections of actual stiction periods are marked. The detection is overly sensitive to minor fluctuations in the PV and OP signals, leading to broken labeling that does not fully cover the continuous stiction band.

In contrast, Figure 13(b) shows the results $n = 24$ consecutive windows are applied. Here, the red-highlighted detection bands are longer and more continuous, closely covering the entire stiction period observed in the process data. This setting captures the complete deadband asymmetry and stick–slip cycles characteristic of valve stiction, with fewer false triggers from process noise. Other window configurations, such as values below 15 or between 15 and 24, produced highly fragmented labeling, while very large $n$ values smoothed the slope ratios excessively, driving $\beta$ toward zero and causing short stiction periods to be overlooked, as the model failed to distinguish them from normal variations. Overall, $n = 24$ provided the best balance between coverage and sensitivity in this study.

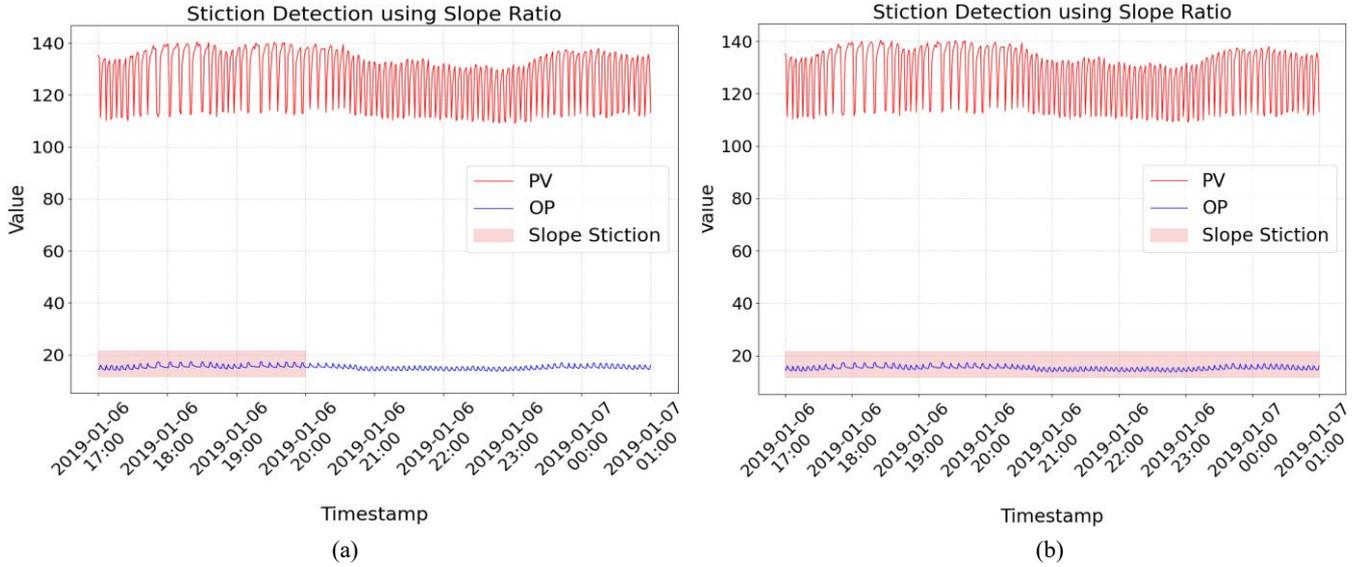

**Figure 13:** Slope Ratio Method labeling using a 60-minute base window. The red-highlighted band indicate stiction periods.
(a) $n = 15$ consecutive windows, producing fragmented and not cover whole stiction band labeling.
(b) $n = 24$ consecutive windows, producing stable labeling cover period aligned with stiction behavior.

The second method, Hotelling's $T^2$, was evaluated using two statistical confidence thresholds, 90% and 60%, selected based on the approach used by Yazdi et al. [2]. Figure 14(a) presents the 90% confidence threshold configuration, which produced sparse detections and failed to consistently identify prolonged stiction bands, missing several true stiction events, whereas Figure 14(b) illustrates the 60% confidence threshold, which generated more frequent detections but suffered from noisy labeling and poor alignment with actual stiction behavior observed in the process data. Intermediate confidence levels, such as 70% or 80%, were also examined during testing. These values yielded performance between the two extremes, providing slightly better coverage than 90% but still prone to false positives and fragmented labeling, similar to 60%, whereas thresholds below 50% barely detected any stiction at all.

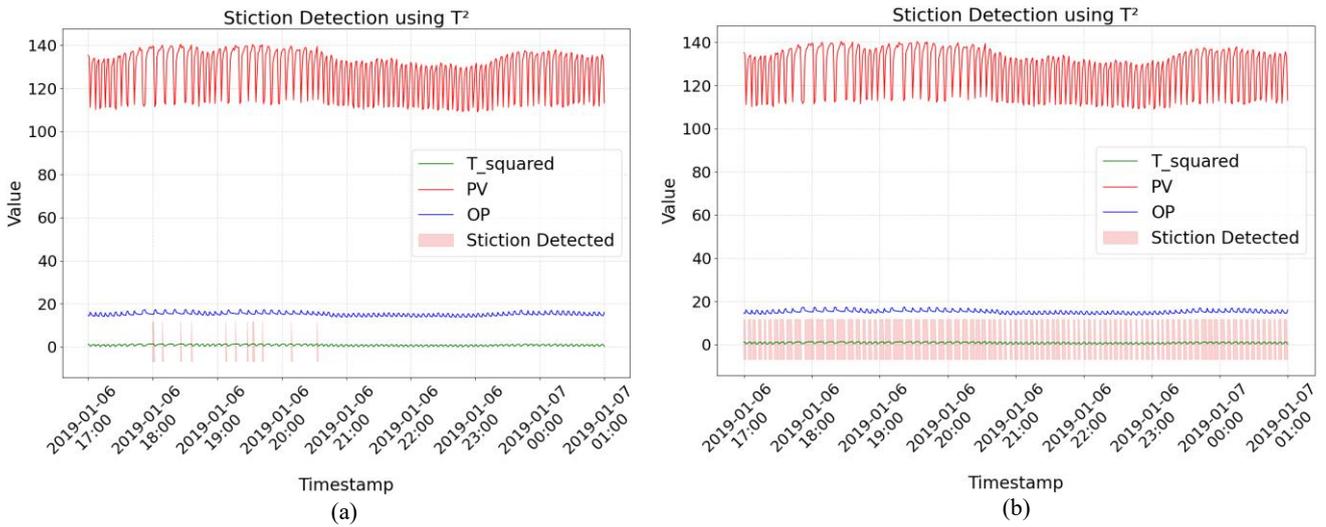

**Figure 14:** Hotelling's T² detection using different confidence thresholds for stiction detection.
(a) 90% confidence limit, resulting in minimal stiction indications.
(b) 60% confidence limit, producing frequent but inconsistent stiction indications.

When compared side-by-side, the Hotelling's T² method, regardless of the confidence threshold applied, showed lower consistency and poorer alignment with actual stiction patterns compared to the Slope Ratio Method. Consequently, the Slope Ratio configuration with a 60-minute base window and 24 consecutive windows was retained as the definitive labeling approach for generating ground truth data, forming the basis for all subsequent model training and evaluation.

## 4.2 Control Valve Stiction Model Evaluation

To evaluate real-time stiction detection performance, the labeled dataset from the slope ratio method using a 60-minute window and 24 consecutive windows was used to train three architectures: CNN, CNN-SVM, and LSTM.

### 4.2.2 Real-Time Stiction Detection

The ability of three models, CNN, CNN-SVM, and LSTM, to correctly identify stiction events as they occurred in the process data was evaluated based on the dataset containing a total of 525,600 labelled events, with 282,240 stiction instances and 243,360 non-stiction instances. The assessment was based on their confusion matrices and performance metrics, focusing on both the correct detection of stiction cases and the accurate recognition of normal operation. All three models achieved perfect recall for stiction, meaning that every stiction instance present in the dataset was detected. However, their ability to correctly identify non-stiction cases, which is critical for avoiding false alarms and unnecessary maintenance actions, varied considerably.

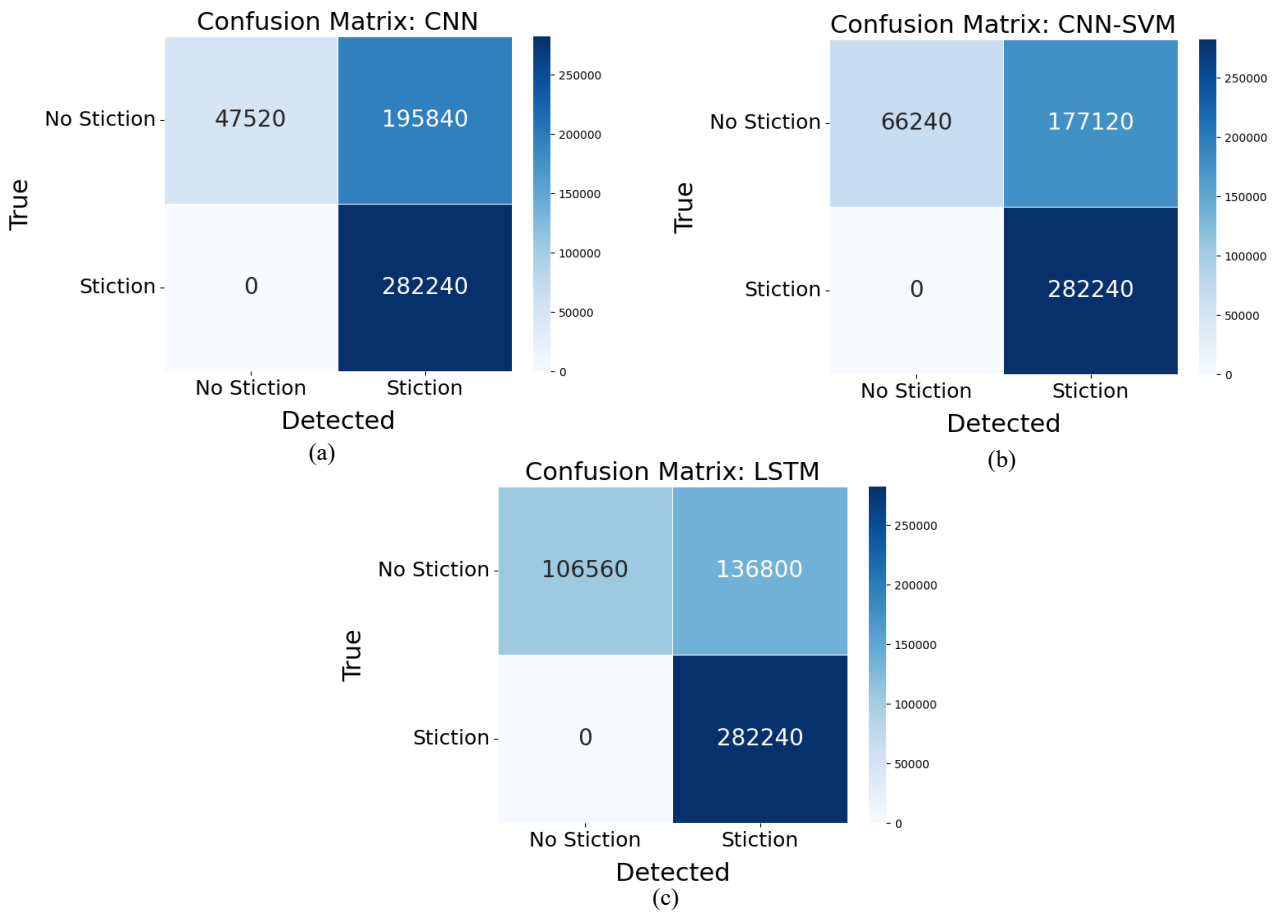

**Figure 15:** Confusion matrices for real-time control valve stiction detection using (a) CNN, (b) CNN–SVM, and (c) LSTM models. The matrices compare predicted labels with slope ratio labelling for both stiction and non-stiction cases, showing classification performance across the three deep learning approaches.

The results in Figure 15 show how the three models perform. In Figure 15(a), the CNN model detects all stiction cases without missing any, which means it is very sensitive to stiction. However, it struggles to correctly identify normal operation, with only 47,520 non-stiction cases classified correctly and 195,840 misclassified. This shows that the model often mistakes normal variations in the signals for faults, which in real situations could lead to unnecessary maintenance.

When the SVM layer is added to create the CNN-SVM model, shown in Figure 15(b), the results start to improve. Like the original CNN, it still detects all stiction cases perfectly. However, it is better at recognizing normal operation, correctly identifying 66,240 non-stiction cases. The SVM layer helps the model make clearer decisions, which reduces false alarms and makes it less likely to mistake normal behavior for faults. Even with this improvement, the model still misclassified 177,120 non-stiction cases as stiction, indicating that further tuning of the SVM kernel parameters or adjusting class balancing in the training data would be necessary to reduce these errors.

The LSTM model in Figure 15(c) gives the best overall performance. It also detects all stiction cases perfectly, but it achieves the highest accuracy for normal operation, misclassifying only 136,800 non-stiction cases. LSTM works well because it analyses how the signals change over time, which helps it tell the difference between real faults and normal changes in the process. This gives a better balance between finding faults and avoiding false alarms, making it the most suitable choice for real-time stiction monitoring.

**Table 4:** Real Time Stiction Detection Performance Comparison

| Metric | CNN | CNN-SVM | LSTM |
|---|---|---|---|
| **Overall** | | | |
| Accuracy | 0.63 | 0.66 | 0.74 |
| Macro avg precision | 0.80 | 0.81 | 0.84 |
| Macro avg recall | 0.60 | 0.64 | 0.72 |
| Macro avg f1-score | 0.53 | 0.59 | 0.71 |
| Weighted avg precision | 0.78 | 0.79 | 0.82 |
| Weighted avg recall | 0.63 | 0.66 | 0.74 |
| Weighted avg f1-score | 0.55 | 0.61 | 0.71 |
| **Stiction** | | | |
| Precision | 0.59 | 0.61 | 0.67 |
| Recall | 1 | 1 | 1 |
| F1-score | 0.74 | 0.76 | 0.8 |
| Support | 282240 | 282240 | 282240 |
| **Non-Stiction** | | | |
| Precision | 1 | 1 | 1 |
| Recall | 0.20 | 0.27 | 0.44 |
| F1-score | 0.33 | 0.43 | 0.61 |
| Support | 243360 | 243360 | 243360 |

In terms of overall performance, as shown in Table 4, all three models demonstrated the ability to detect stiction events in real time, but with varying degrees of accuracy. The CNN model achieved an overall accuracy of 0.63, while CNN-SVM showed a modest improvement at 0.66. The LSTM model delivered the highest overall accuracy at 0.74, reflecting a stronger balance between detecting true faults and avoiding false alarms.

For stiction detection, all models achieved a perfect recall of 1.0, meaning that every stiction event in the dataset was successfully identified. This ensures that no actual stiction occurrence went undetected. However, the precision values reveal some differences. CNN achieved a precision of 0.59, CNN-SVM slightly improved to 0.61, and LSTM performed best at 0.67. Higher precision indicates fewer false positives within the detected stiction events, making LSTM the most accurate in confirming true faults.

When focusing on non-stiction detection, the differences become more pronounced. While all models recorded perfect precision of 1.0, their recall values, which reflect the ability to correctly identify normal operating conditions, were notably lower. CNN had the lowest non-stiction recall at 0.20, leading to frequent false alarms. CNN-SVM improved slightly to 0.27, while LSTM achieved a significantly better recall of 0.44. This indicates that LSTM was the most effective at recognizing normal conditions without compromising its ability to detect stiction, making it the most balanced model overall.

*4.2.3 Early Stiction Prediction*
The early stiction prediction task evaluated the ability of the CNN, CNN-SVM, and LSTM models to forecast stiction events before they occurred, using the same dataset of 525,600 labelled events applied in the real-time detection task. In Figure 16(a) to Figure 16(c), the heatmaps illustrate the accuracy of each model for different combinations of input and lookahead windows. The x-axis of the heatmaps represents the input window, which is the number of consecutive past data segments used by the model to make a prediction, while the y-axis represents the lookahead window, which indicates how far into the future the prediction is made. Each window corresponds to one hour of process data, matching the 60-minute segment used in the slope ratio method for stiction labeling. This alignment ensures that the input features and the labeled events are based on the same time resolution, allowing consistent detection and prediction comparisons. By systematically varying both the input and lookahead windows, the models were evaluated to determine which configurations delivered the highest predictive accuracy.

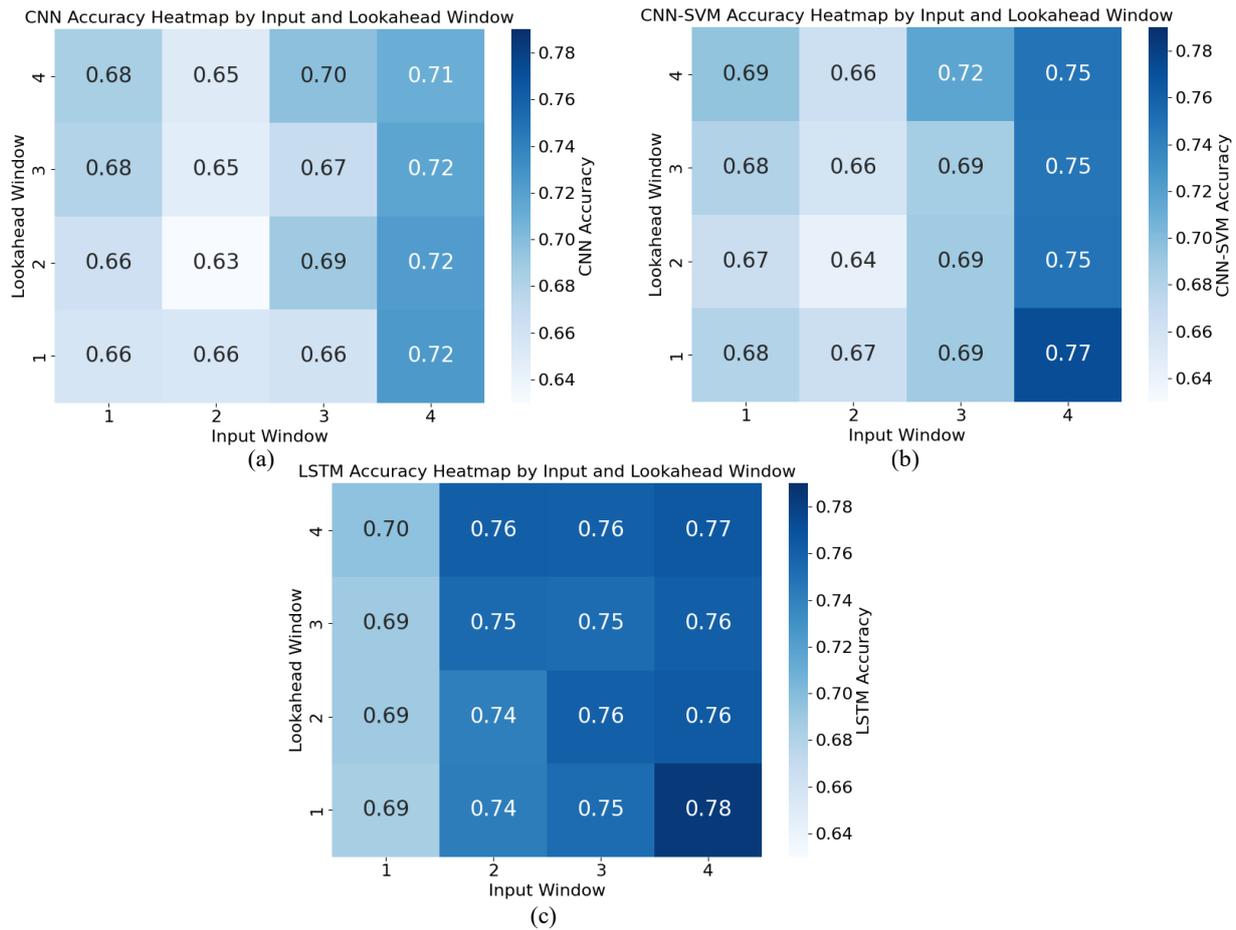

**Figure 16:** Accuracy heatmaps for early stiction detection using different input and lookahead window configurations for (a) CNN, (b) CNN–SVM, and (c) LSTM models. Each cell represents model accuracy for the corresponding combination of input and lookahead windows.

The CNN model, presented in Figure 16(a), demonstrates a steady improvement in predictive accuracy as the length of the input window increases. Its highest accuracy, 0.72, is achieved with a four-hour input window, and this performance remains consistent irrespective of whether the prediction horizon is one, two, three, or four hours. This stability suggests that the CNN benefits from a longer historical context, enabling more effective identification of stiction patterns. In contrast, shorter input windows, particularly one or two hours, yield reduced accuracy, indicating that the model's ability to detect temporal patterns diminishes with limited historical data.

The introduction of an SVM classification layer in the CNN-SVM model, shown in Figure 16(b), results in an overall enhancement in performance relative to the baseline CNN. The highest accuracy recorded is 0.77, achieved with a four-hour input window and a one-hour prediction horizon, illustrating strong short-term predictive capability when supported by richer historical input. Although performance remains robust for extended prediction horizons, a slight decrease in accuracy is observed as the forecast window increases. Nonetheless, across comparable configurations, the CNN-SVM consistently outperforms the CNN, underscoring the value of combining CNN-based feature extraction with the classification precision of SVM.

The LSTM model, depicted in Figure 16(c) exhibits superior performance across nearly all configurations, achieving a peak accuracy of 0.78 with a four-hour input window and a one-hour prediction horizon. Notably, the LSTM maintains accuracy levels above 0.75 for longer prediction horizons, reflecting its inherent capability to capture and utilize long-term temporal dependencies within the process data. This sequential memory structure enables the LSTM to more effectively distinguish between genuine stiction patterns and normal process variability, providing a distinct advantage for early and reliable stiction prediction in industrial applications.

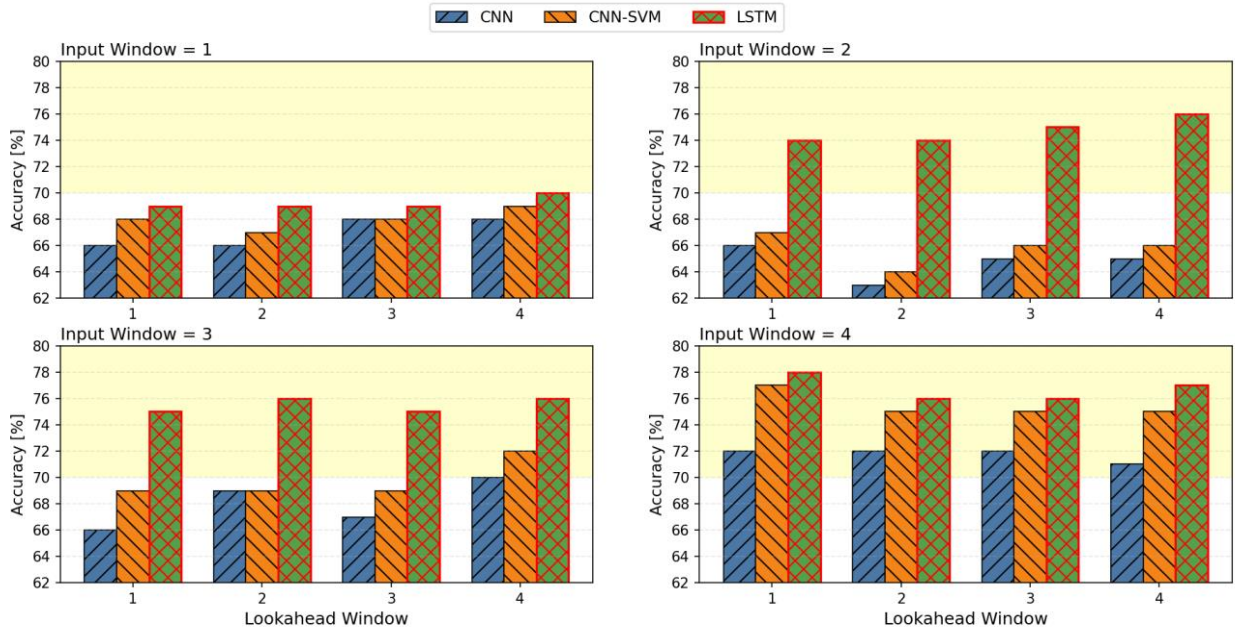

**Figure 17:** Accuracy comparison of CNN, CNN-SVM, and LSTM models across input–lookahead window combinations for early stiction prediction. The yellow-shaded region denotes the project's target threshold of 70 percent accuracy. LSTM consistently exceeds this benchmark across most settings, while CNN-SVM meets it in several cases and CNN only at larger input windows.

Figure 17 compares the performance of CNN, CNN-SVM, and LSTM across different input–lookahead window settings. At the smallest setting, Input Window = 1, LSTM clearly outperforms the others, consistently exceeding the 70 percent (0.70) benchmark, while CNN and CNN-SVM remain below, highlighting their limitations in capturing short-term dynamics. When the input window increases to 2, the gap widens: LSTM stabilizes above 74 percent, whereas CNN-SVM fluctuates around the threshold and CNN lags further behind. With Input Window = 3, the contrast becomes more striking as LSTM sustains strong accuracy above 74 percent, CNN-SVM achieves marginal gains around 70 percent, and CNN struggles to keep pace. Finally, with input window 4, the models diverge most strongly. CNN only peaks at 72 percent, CNN-SVM improves to 77 percent in the 4–1 setting but falls off afterwards, while LSTM proves most reliable, maintaining accuracies above 75 percent and reaching its highest at 78 percent. Overall, the results highlight LSTM's robustness in modeling temporal dependencies, CNN-SVM's partial but limited advantage over CNN, and the general trend that longer input windows with shorter horizons yield stronger predictions.

LSTM's advantage also extends to its ability to accurately identify non-stiction periods, thereby reducing false positives while still capturing stiction events with high recall. This behavior aligns with the real-time detection outcomes presented in Appendix 9.1, where LSTM demonstrated higher recall for non-stiction classification compared to CNN and CNN-SVM. The improved balance between precision and recall enhances overall predictive accuracy, supporting LSTM's suitability for practical implementation in industrial systems, where minimizing unnecessary maintenance alerts is as critical as detecting stiction events early.

# 5. DISCUSSION

*5.1 Key Insights*

The LSTM model in this study consistently outperformed CNN and CNN-SVM in both real-time detection and early prediction tasks. For real-time detection, it achieved the highest overall accuracy of 74 percent, correctly identifying more non-stiction cases than the other models. For early predictions, it reached a maximum accuracy of 78.34 percent for a one-hour forecast horizon. These results confirm the strength of LSTM architectures in capturing temporal dependencies, consistent with other industrial anomaly detection and fault prediction studies, such as Malhotra et al. [14] with 83 percent accuracy, Filonov et al. [15] with 87 percent, and Shokouhmand et al. [16] reaching 93 percent. Unlike prior LSTM applications that often focused on rotating machinery or generic process faults, this work applies LSTM to refinery control valve stiction using real OP and PV data from plant historians, addressing a gap where most earlier studies relied on simulated or non-valve datasets.

While CNN-SVM improved over CNN in balancing stiction and non-stiction classification, it still lagged behind LSTM in predictive performance. This mirrors limitations seen in CNN–PCA hybrids [10], where dimensionality reduction aids classification but temporal generalization remains constrained. Compared with CNN-based methods that of Henry et al. [8] (90 percent on simulated data) and Gunnell et al. [9] (75.76 percent with real OP and PV), the models in this work demonstrate competitive detection performance while uniquely extending capability to multi-step forecasting. Furthermore, unlike previous CNN-focused research that primarily addressed single-time detection, the present study evaluates varying input and lookahead windows, showing how shorter horizons yield higher accuracy but less intervention time, whereas longer horizons extend operator reaction windows with a moderate drop in accuracy.

Overall, this study provides the first direct benchmark between CNN, CNN-SVM, and LSTM for both real-time stiction detection and early prediction under identical industrial datasets and labeling conditions. The results show that the LSTM approach offers a practical pathway for integrating predictive valve diagnostics into refinery maintenance strategies, advancing beyond the purely reactive detection frameworks that dominate current literature.

*5.2 Practical Implications*

In operational environments, the LSTM model with an input window of four and a lookahead window of one yielded the best performance with 78.34 percent accuracy. This configuration is suitable for processes that require rapid fault detection but have limited intervention time. When the lookahead window was extended to four intervals, the LSTM model still achieved 76.5 percent accuracy and provided approximately four hours of advance warning. This level of forecasting is sufficient for most industrial scenarios where scheduled intervention is necessary to prevent performance degradation [26].

Beyond model performance, integrating this framework into a plant's control system architecture offers significant potential. The predictive model can be deployed as a module within existing distributed control systems or connected through communication interfaces to supervisory control platforms. It can continuously receive real-time PV and OP data from the plant historian, process these through the trained model, and deliver advisory alerts directly to the human–machine interface. Such integration would allow operators to act on early warnings without interrupting ongoing control, and the system could be extended to trigger maintenance work orders automatically in the plant's computerized maintenance management system.

Economically, the findings suggest substantial cost savings compared with full valve replacement programs. For example, replacing 11,232 conventional control valves in one oil and gas refinery with smart positioners would cost over $28 million USD, assuming a per-unit price of roughly $2,500 USD based on current market rates for digital positioners [27]. These prices are corroborated by supplier data indicating that digital models typically range from $2,500 to $6,000 USD per unit depending on features and diagnostic capabilities [28]. In contrast, engaging predictive maintenance diagnostics allows targeted interventions on valves showing early signs of stiction, avoiding unnecessary capital expenditure. Beyond equipment costs, this approach also reduces manual inspections, minimizes compressed air use, lowers energy waste, and prevents unplanned downtime, safeguarding throughput and product quality while improving cost efficiency.

*5.3 Limitations*

There are several limitations that must be considered. First, the dataset labeling relied on slope ratio analysis rather than expert annotation, which may have introduced labeling inaccuracies. Similar challenges have been documented in studies that use data driven labeling methods for neural network training [23], [25]. Second, the model's performance decreased when applied to a different valve not used during training. For instance, the accuracy dropped from 78.34 percent on Control Valve A to 73 percent on Control Valve B. This suggests that model generalization remains limited and may require equipment specific calibration, a challenge also noted in recent industrial fault detection literature [26]. Third, the study's validation was conducted using valves operating under similar process conditions. Broader validation across a wider range of valve types and process scenarios is necessary to ensure transferability. Finally, CNN models produced high rates of false positives that could lead to operator fatigue and diminish system credibility unless complemented by additional filtering mechanisms or advisory decision support tools.

# 6. CONCLUSIONS

*6.1 Main Findings and Contribution*

This study explored the application of deep learning models, specifically CNN, CNN-SVM, and LSTM, into detecting and forecasting control valve stiction in industrial process systems. All three models demonstrated reliable real-time detection and, more importantly, enabled an early prediction framework through sliding window configurations, showing that stiction events can be forecast hours in advance using refinery OP and PV data.

To the best of the authors' knowledge, this work is the first to develop and validate an early prediction framework for control valve stiction using real industrial datasets rather than simulations or proxy systems. This contribution establishes a practical foundation for predictive diagnostics in control loops of continuous process industries such as refineries, where condition-based maintenance can be achieved without requiring costly hardware upgrades.

Beyond technical feasibility, the framework illustrates how data-driven approaches can reduce downtime, optimize energy efficiency, and improve decision-making in complex industrial control environments. These findings highlight the potential for integrating predictive stiction monitoring directly into industrial control architectures, supporting more proactive and economically efficient maintenance strategies.

*6.2 Recommendations for Future Work*

To enhance the robustness and applicability of the proposed system, future work should focus on improving the accuracy of ground truth labeling by incorporating expert-verified annotations or by combining multiple detection techniques. Expanding the dataset to include a wider variety of control valve types, sizes, and process conditions would improve model generalization across diverse operational contexts. Further optimization of the deep learning models through hyperparameter tuning and the application of interpretable artificial intelligence methods can increase both performance and operator trust [29]. For example, adjusting parameters such as the learning rate, number of hidden layers, number of neurons per layer, batch size, or dropout rate during training can significantly influence model accuracy and generalization. Developing a real-time advisory system that integrates statistical techniques such as the cumulative sum method could enhance the practicality of alerts and reduce false positives [30], [31]. Pilot testing in live industrial environments using actual distributed control or supervisory systems will be essential to validate integration and assess performance. Alternatively, deployment within a Digital Twin could allow safe testing of valve behavior under various scenarios, enabling risk-free prototyping, tuning, and operator training before full implementation. A techno-economic analysis comparing predictive maintenance with traditional hardware replacement can further quantify cost savings and encourage wider adoption of intelligent diagnostic tools in the process industry.

# 7. ACKNOWLEDGEMENT

Natthapong Promsricha sincerely thanks Thaioil Public Company Limited for providing operational data, technical insight, and research sponsorship that made this study possible. Gratitude is also extended to the Department of Mechanical Engineering, University College London, for its academic guidance, research resources, and supportive environment that enabled the successful completion of this work.

# 9. APPENDIX

*9.1* Time-Series Comparison of CNN, CNN-SVM, and LSTM Predictions for Real-Time Control Valve Stiction Detection over a One-Year Dataset.

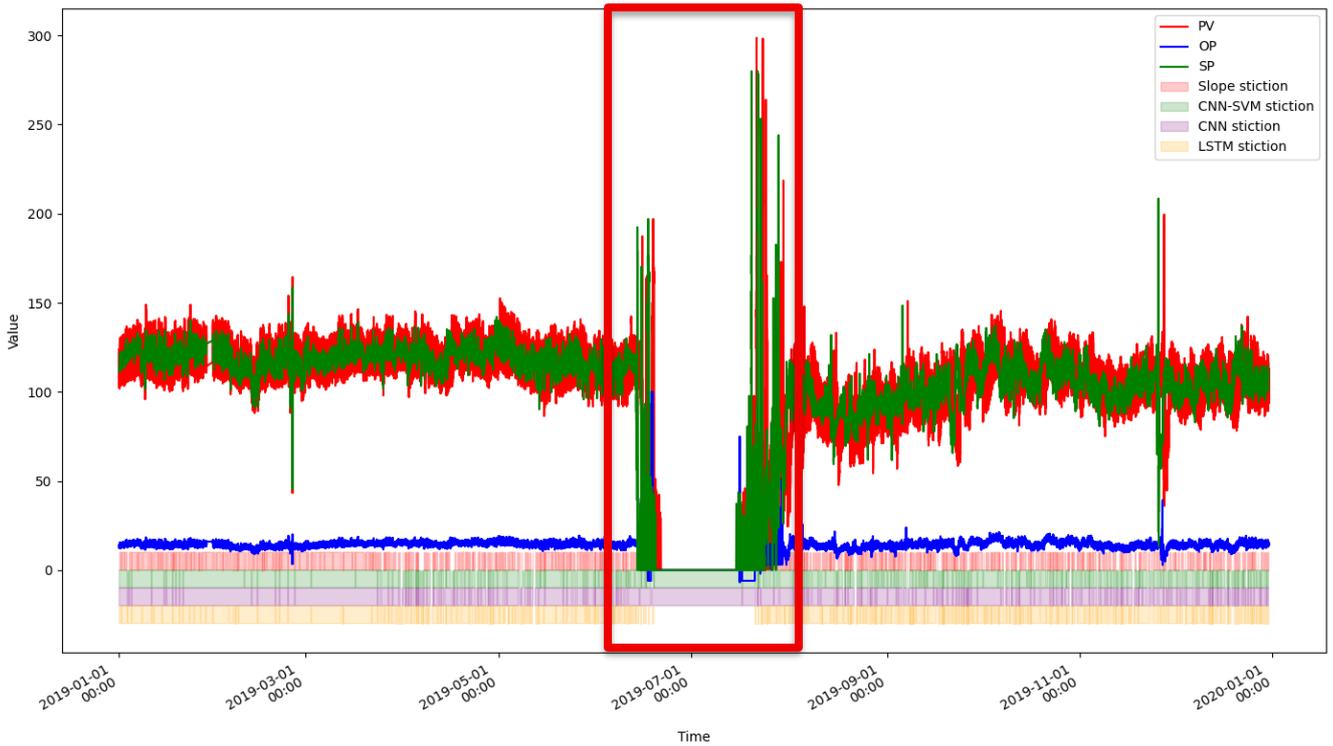

*9.2* Table paring input and lookahead window with graph for illustration.

| Detect Window (Learning Input) | Lookahead Window (Prediction Output) |
|---|---|
| 1 | 1 |
| 1 | 2 |
| 1 | 3 |
| 1 | 4 |
| 2 | 1 |
| 2 | 2 |
| 2 | 3 |
| 2 | 4 |
| 3 | 1 |
| 3 | 2 |
| 3 | 3 |
| 3 | 4 |
| 4 | 1 |
| 4 | 2 |
| 4 | 3 |
| 4 | 4 |

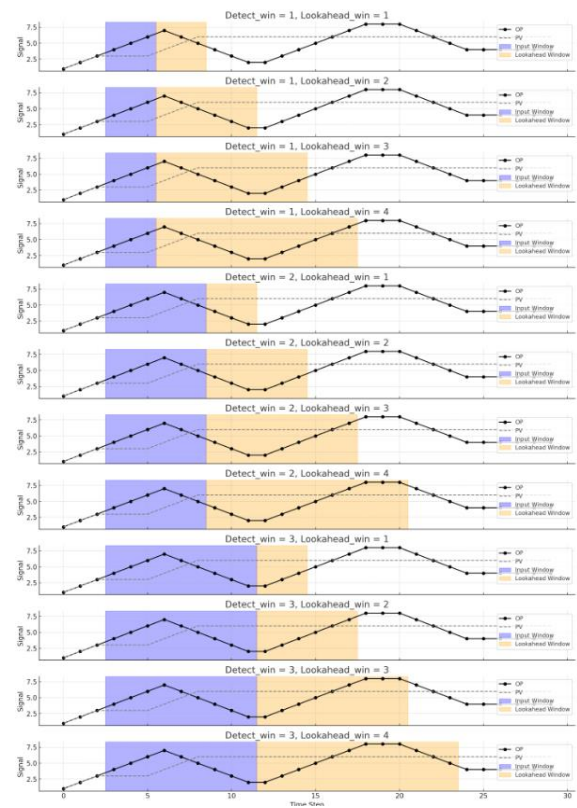